\documentclass[11pt,x11names]{article}
\usepackage{setspace}
\usepackage{misc/emnlp2023}

\usepackage{times}
\usepackage{xcolor}
\usepackage{graphicx}
\usepackage[T1]{fontenc}
\usepackage[utf8]{inputenc}
\usepackage{microtype}
\usepackage{booktabs}
\usepackage{todonotes}
\usepackage{amsmath,amssymb}
\usepackage{cleveref}
\usepackage{xspace}
\usepackage{soul} %
\usepackage{caption}
\usepackage{multirow}
\usepackage{ulem}
\usepackage{float}
\usepackage{array}
\usepackage{bm}
\usepackage{xfrac}
\usepackage{tcolorbox} %
\usepackage[shortlabels]{enumitem}
\usepackage[outline]{contour}%
\usepackage{tikz}
\usepackage{algorithm,algpseudocode,algorithmicx}

\crefname{algorithmB}{algorithm}{algorithms}
\Crefname{algorithmB}{Algorithm}{Algorithms}
\DeclareCaptionType{algorithmB}[Algorithm][List of algorithms]

\setlist[itemize]{noitemsep,left=0mm}

\usepackage{tabularx}

\crefname{lstlisting}{listing}{listings}
\Crefname{lstlisting}{Listing}{Listings}
\crefname{equ}{equation}{equations}
\Crefname{equ}{Equation}{Equations}
\Crefname{algorithm}{Algorithm}{Algorithms}
\crefname{example}{example}{examples}
\Crefname{example}{Example}{Examples}
\crefname{prompt}{prompt}{prompts}
\Crefname{prompt}{Prompt}{Prompts}
\DeclareCaptionType{example}[Example][List of examples]
\DeclareCaptionType{prompt}[Prompt][List of prompts]
\DeclareCaptionType{equ}[Equation][List of equations]

\usepackage{marginnote}

\definecolor{TodoColor}{rgb}{1,0.7,0.6}
\definecolor{TodoColor2}{rgb}{0.7,0.7,0.9}
\definecolor{TodoColor3}{rgb}{0.5,0.8,0.5}

\newcommand{\cometkiwi}{CometKiwi\xspace}

\makeatletter\def\Hy@Warning#1{}\makeatother
\let\svthefootnote\thefootnote
\newcommand\blankfootnote[1]{%
  \let\thefootnote\relax\footnotetext{#1}%
  \let\thefootnote\svthefootnote%
}

\graphicspath{{img/}}

\title{
    A Bayesian Optimization Approach to Machine Translation Reranking
}

\author{
    Julius Cheng$^1$\quad
    Maike Züfle$^2$\quad
    Vilém Zouhar$^3$\quad
    Andreas Vlachos$^1$ \\[1em]
    $^1$University of Cambridge \quad
    $^2$Karlsruhe Institute of Technology \quad
    $^3$ETH Zürich\\
    \texttt{\{\href{mailto:jncc3@cam.ack.uk}{\color{black} jncc3},\href{mailto:av308@cam.ack.uk}{\color{black} av308}\}@cam.ac.uk} \quad
    \texttt{\href{mailto:maike.zuefle@kit.edu}{\color{black} maike.zuefle@kit.edu}} \quad
    \texttt{\href{mailto:vzouhar@ethz.ch}{\color{black} vzouhar@ethz.ch}}
}

\begin{document}

\onecolumn

{
\setlength{\parindent}{0cm}
}

\clearpage

\twocolumn

\maketitle

\maketitle

%
%

\newcommand{\candsobs}{\ensuremath{\mathcal{C}_\mathrm{obs}}}
\newcommand{\candsobsbar}{\ensuremath{\bar{\mathcal{C}}_\mathrm{obs}}}
\newcommand{\cands}{\ensuremath{\mathcal{C}}}

\begin{abstract}

\textit{Reranking}, or scoring a list of prediction candidates from a machine translation system with an external scoring model and returning the highest-scoring candidate, remains a simple and effective method for improving prediction quality.
However, reranking with high quality scoring models can add substantial computational cost to the translation pipeline, which we address in this work by framing list reranking as a Bayesian optimization (BayesOpt) problem over the candidate list, where unknown scores are modeled with a Gaussian process.
This algorithm scores candidates iteratively, choosing next candidates by balancing between exploration, choosing to score those that differ from candidates already scored, and exploitation, choosing to score those that resemble high-scoring candidates.
This procedure finds high-scoring candidates while scoring only a fraction of the candidates list; given candidate lists of 200 random samples (before deduplication), our method achieves the same \cometkiwi score using only 70 scoring evaluations on average compared to scoring a random subset of 180 candidates.
We also propose multi-fidelity BayesOpt for list reranking, where scores obtained from a noisier but cheaper \textit{proxy} scoring model are incorporated into the search process.
We show that well-trained distilled proxy scorers can further improve the performance of BayesOpt.
\end{abstract}

\blankfootnote{Code repository: \href{https://github.com/juliusc/bayesopt_reranking}{github.com/juliusc/bayesopt\_reranking}}

\section{Introduction}
\label{sec:introduction}

\begin{figure}[htbp]
\includegraphics[width=\linewidth]{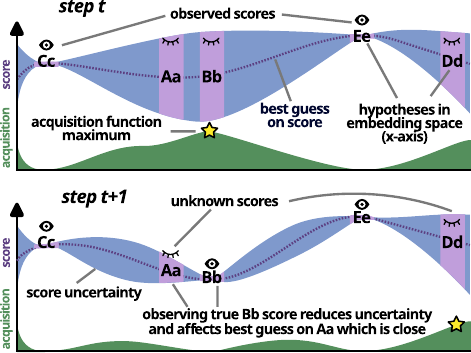}
\caption{A machine translation system generates candidates \texttt{Aa}, \texttt{Bb}, \texttt{Cc}, \texttt{Dd}, and \texttt{Ee}.
The goal of BayesOpt is to find the highest scoring candidate with fewer scoring calls.
An acquisition function selects the next candidate to score repeatedly until budget is reached, and the candidate with the highest score so far is returned.
}
\label{fig:highlevel_schema}

\end{figure}

\textit{Reranking} is a framework for prediction where probabilistic \textit{generator} model produces a list of candidates, and a separate \textit{evaluator} or \textit{scoring model} produces scores for each of the candidates which are use to determine the final prediction. Reranking has a long history in natural language processing for sequential prediction problems such as dependency parsing \citep{collins-koo-2005-discriminative,charniak-johnson-2005-coarse} and language modeling problems such as summarization \citep{ravaut-etal-2022-summareranker} and machine translation \citep[MT;][]{fernandes-etal-2022-quality}. 


The quality of models for automatic MT evaluation has surged in recent years due to innovations in neural network architecture \citep{rei-etal-2020-comet,juraska-etal-2023-metricx,sellam-etal-2020-bleurt} as well as the abundance of training data \citep{freitag-etal-2023-results,kocmi-etal-2024-findings}. These evaluation models are often repurposed for reranking to further improve the performance of an MT system.
For instance, in the WMT 2024 shared task \citep{kocmi-etal-2024-findings}, 5 out of 19 systems, including the overall best submission \citep{rei-etal-2024-tower} use reranking with Comet models \citep{rei-etal-2020-comet} and/or minimum Bayes risk decoding \citep[MBR;][]{eikema-aziz-2020-map}, which can be interpreted as a form of reranking. Prior to the application of automatic evaluation metrics to reranking, other scoring methods have been proposed, including discriminatively trained classifiers \citep{lee-etal-2021-discriminative,bhattacharyya-etal-2021-energy} and noisy channel decoding \citep{yee-etal-2019-simple}.

So, while LMs for MT generation for greatly improved in recent years, scoring models have seen a commensurate increase in quality \citep{zerva-etal-2022-findings}, and thus reranking remains relevant method for improving translation quality. However, the scoring models have also grown dramatically in size, increasing the computational requirements for reranking.




In this work, we address the computational cost of reranking by framing it as a search problem over the list of candidates. The goal of search in this setting is to find high-scoring candidates in a small number of steps, thereby avoiding the cost of scoring the full list. Our proposed algorithm uses Gaussian processes to model uncertainty about unseen scores and Bayesian optimization \citep[BayesOpt;][]{Shahriari2016TakingTH} to choose which candidates to score next.

GPs are flexible priors over functions which are able to model the complex and nonlinear relationship between each candidate and its score. GPs make very few assumptions about the distribution and base their predictions are mostly on observed points, which enables them to easily adapt to different candidate lists across translation instances. BayesOpt is a sequential black-box optimization method that uses the posterior mean and variance of unobserved data points to decide which points to evaluate next.

We apply BayesOpt and GPs (BayesOpt+GP) to MT list reranking in a straightforward manner and show that it obtains close to the maximum achievable score with only a fraction of score evaluations.
For example, the maximal obtainable score across 200 randomly sampled candidates on our test set is 0.8216 \cometkiwi; our method achieves 0.8210 with 70 score evaluations on average, while scoring 70 random candidates attains 0.8149, a difference of 0.0061 which is likely to be human-detectable according to \citep{kocmi-etal-2024-navigating}.
We also propose a number of search-based baselines which outperform random selection, all of
which are outperformed by BayesOpt+GP.

Then, building upon previous works that use a faster but noisier proxy scoring function to prune the candidate list \citep{fernandes-etal-2022-quality,eikema-aziz-2022-sampling}, we propose a \textit{multi-fidelity} extension to BayesOpt which incorporates proxy scores to improve estimation.
This is related in motivation to coarse-to-fine methods \citep{coarse-to-fine} and model cascading \citep{chen2023frugalgptuselargelanguage}, where the use of a faster proxy model reduces the use of the main model.
In our multi-fidelity experiments, we find that smaller proxy scoring models distilled from the main model can assist BayesOpt+GP in finding high-scoring candidates earlier.

\section{Background}

\subsection{Translation generation and reranking}

In a typical machine translation setting, a conditional language model (LM) is trained to model the probability of the next token $y_t$ given a source sentence $x$ and previous tokens: $p(y_t|x,y_1,...,y_{t-1})$.
These probabilities can be autoregressively combined to model a sequence probability $p(y|x)$.
Usually, beam search is used to search for a $y$ which maximizes log probability combined with a length normalization objective \citep{wu2016googlesneuralmachinetranslation}.

In a basic list reranking setting, given $x$, the LM is used to generate a candidate list $\mathcal{C}_x=[y_1,...,y_n]$ with a \textit{decoding algorithm} such as beam search or ancestral sampling.
A scoring function $s(x,y_i)$ is then applied to each $y_i\in\mathcal{C}_x$, and the best scoring sequence $\arg\max_{y_i\in\mathcal{C}_x} s(x,y_i)$ is returned.
A common choice of scoring function is a quality estimation (QE) model which directly predicts a scalar value representing the quality.

Reranking with high-quality evaluation metrics has been shown to be highly effective at improving translation output \citep{freitag-etal-2022-high}, though it can skew results when the same metric is also used for evaluation \citep{kocmi-etal-2024-findings}.
Reranking performance improves as the number of candidates increases \citep{vernikos-popescu-belis-2024-dont} and when multiple scoring metrics are combined to form a stronger prediction \citep{fernandes-etal-2022-quality}. 

Reranking adds significant computational costs to prediction and may be prohibitive to use at test time, but it can be used to benefit LM training instead of test time prediction; high-quality predictions obtained from reranking can be used for knowledge distillation \citep{wang2024dontthrowawaydata} and self-training \citep{finkelstein2024mbrqefinetuningtrainingtime}. Such methods can improve the performance of an MT system without additional costs during test time.

Previous work on efficient reranking for MT is relatively limited. \citet{fernandes-etal-2022-quality} and \citet{eikema-aziz-2022-sampling} perform a two-stage reranking by first pruning with a faster and noisier scoring function to a fixed size before evaluating the target score. There has been recent interest in efficient approximations for MBR \citep{cheng-vlachos-2023-faster,deguchi-etal-2024-centroid,trabelsi2024efficientminimumbayesrisk,vamvas-sennrich-2024-linear}, but these methods are not applicable to general scoring functions. \citep{singhal-etal-2023-eel} propose to represent the candidate space compactly in a lattice over which a token-level reranker can efficiently score many candidates. In this work, we attempt to address a more general setting: the reranking of candidate lists with arbitrary black-box scoring functions.

\subsection{Bayesian optimization with\linebreak Gaussian process prior}
\label{sec:bayesopt_gp}

Bayesian optimization is a sequential algorithm for optimizing a black-box function $f$. $f$ is assumed to be drawn from a prior distribution over functions.
The main loop of BayesOpt is as follows: given a set of (possibly noisy) observations of $f(a_1),...,f(a_i)$, the prior distribution over $f$ is updated to a posterior distribution with Bayes theorem.
An \textit{acquisition function} determines a \textit{query point} $a_{i+1}$ at which to evaluate $f$ next.
$f(a_{i+1})$ is evaluated and added to the set of observations.
This repeats until a stopping criteria is reached.
The principal design choices in BayesOpt are the prior distribution of $f$ and the acquisition function.

A common choice of prior is the Gaussian process, which assumes that any subset of points $f(a_1),...,f(a_i)$ are drawn jointly from a multivariate Gaussian distribution $\mathcal{N}(\mu, \mathcal{K})$, where $\mathcal{K}$ is the covariance matrix defined by a kernel function such as the radial basis function kernel (RBF).
RBFs define the covariance of two points $a$ and $a'$ as:
\begin{align}
\label{eq:rbf}
\mathcal{K}_\mathrm{RBF}(a,a') = \exp \left(-\frac{||a-a'||^2}{2w^2} \right),
\end{align}
where $w$ is the \textit{bandwidth} hyperparameter which determines scaling.
The choice of kernel dictates prior assumptions about the shape of $f$; with RBF, points that are closer in Euclidean space have larger covariance. RBFs are a popular choice of kernel due their ability to adapt to complex nonlinear functions. 

The assumption that $f(a_1),...,f(a_i)$ are jointly Gaussian gives rise to a convenient posterior distribution. Given a vector of observed data points $\textbf{a}$ and their observed values $f(\textbf{a})$, the posterior mean $\mu_a$ and variance $\sigma_a$ of a point $a$ are given by the conditional multivariate Gaussian distribution:
\begin{align}
\mu_a &= \mu+\mathcal{K}(a, \mathbf{a})(\mathcal{K}(\mathbf{a},\mathbf{a}) + \sigma^2 I)^{-1}f(\mathbf{a})  \label{eq:gaussian_posterior_mu}\\
\begin{split}
\sigma_a &= \mathcal{K}(a,a) + \sigma^2 - \\
 & \qquad \mathcal{K}(a, \mathbf{a})(\mathcal{K}(\mathbf{a},\mathbf{a}) + \sigma^2 I)^{-1}\mathcal{K}(\mathbf{a}, a) \label{eq:gaussian_posterior_sigma}
\end{split}
\end{align}
where $\mu$ is the unconditional mean of the distribution, $\sigma^2$ is a constant Gaussian noise on observations, $I$ is the identity matrix, and $\mathcal{K}$ here returns elementwise kernel values when given vector arguments.

The acquisition function in BayesOpt is the strategy for selecting the next point to evaluate in the optimization process. Acquisition functions can seek the highest expected improvement \citep[EI;][]{mockus1974bayesian}, an upper confidence bound if the scores are noisy \citep{Srinivas2009InformationTheoreticRB}, or information gain \citep{Hennig2011EntropySF}. We use EI, defined as:
\begin{equation}
    \alpha(a) = \mathbb{E}[\max(f(a) - f(a^+), 0)],
\end{equation}
where $a^+$ is the location of the current best observation. When $f$ is Gaussian and there is no observation noise, this has the following closed-form solution \citep{Jones2001ATO}:
\begin{equation}
\label{eq:ei}
    \alpha(a) = \sigma_a(z\cdot\mathrm{cdf}(z) + \mathrm{pdf}(z)),
\end{equation}
where $z=\frac{f(a^+)-\mu_a}{\sigma_a}$, and $\mathrm{cdf},\mathrm{pdf}$ are the Gaussian cumulative distribution function and probability density function, respectively. EI encourages both exploration of uncertain points and exploitation of high-scoring points; the quantity in Equation \ref{eq:ei} can be increased by increasing $\mu_a$ or $\sigma_a$.

The generality of BayesOpt and modeling freedom enjoyed by GPs make them suitable for a great variety of tasks, including spatial monitoring \cite{krause} and hyperparameter optimisation \cite{bergstra2011hyperopt}. GPs have been applied to text regression tasks \citep{beck-etal-2013-shef,beck-etal-2014-joint,beck-cohn-2017-learning}, but they are not as well-studied in NLP compared to many other domains.

\section{Methods}

\subsection{MT reranking with Bayesian optimization}
\label{sec:bayesopt+gp}

\begin{figure*}[t]
{
\hrule
\vspace{1mm}
\textbf{Inputs}: main metric $s$, proxy metric $s'$, budget $n$ for evaluating $s$, hypotheses $\cands$, number of initial main scores $\alpha$, number of initial proxy scores $\beta$, scoring budget $n$, batch size $k$, precomputed multi-fidelity kernel $\mathcal{K}_{mult}$. \\
\textbf{Output}: hypothesis with the highest observed score $\arg\max_{y\in\candsobs}s(y)$.
\vspace{1mm}
\hrule
\vspace{1mm}

\begin{algorithmic}[1]


\State $\cands'_{\text{obs}} \gets \binom{\cands}{\min(\beta,|\cands|)},\cands_{\text{obs}} \gets \binom{\cands'_{\text{obs}}}{\min(\alpha,|\cands|)}$
\Comment{Sample initial subsets}
\State $S_{\text{obs}} \gets \{ s(y) | y \in \cands_{\text{obs}}\}$ \Comment{Compute scores for main scoring function}
\State $S'_{\text{obs}} \gets \{s'(y) | y \in \cands'_{\text{obs}}\}$ \Comment{Compute proxy scores}

\While{$|\candsobs| < n$ \textbf{and} $|\cands_{\text{obs}}| < |\cands|$}
    \State $\candsobsbar\gets\cands\setminus\candsobs$
    \Comment Get complement of $\candsobs$
    \State $\hat{S}\gets \text{Norm}(S_{\text{obs}}),\hat{S'}\gets \text{Norm}(S'_{\text{obs}})$ \Comment Normalize observed scores to 0 mean, 1 variance
    \State $y_{\text{best}} \gets \arg\max_{y \in \cands_{\text{obs}}} \hat{S}(y)$
    \Comment Get best observed point
    \State $\forall y \in \candsobsbar: \mu_y, \sigma_y \gets$ calculate posterior using $y,\mathcal{K}_\mathrm{mult},\hat{S},\hat{S'}$ \Statex \Comment{GP posterior as in \Cref{eq:gaussian_posterior_mu,eq:gaussian_posterior_sigma}}
    
    \State $\forall y \in \candsobsbar: \gamma_y \gets \text{EI}(y_{\text{best}} , \mu_y, \sigma_y)$  \Comment{Expected improvement as in \Cref{eq:ei}}
    \State $\cands_{\text{top-}k} \gets \arg\mathrm{topk}_{y\in\candsobsbar} \gamma_y$
    \Comment{Select $k$ best hypotheses based on EI}
    \State $S_{\text{obs}} \gets S_{\text{obs}} \cup \{ s(y) | y \in \cands_{\text{top-}k}\}$
    \Comment{Compute scores for selected hypotheses}
    \State $\cands_{\text{obs}} \gets \cands_{\text{obs}} \cup \cands_{\text{top-}k}$ \Comment{Update observed hypotheses}

\EndWhile
\State \Return  $\arg\max_{y \in \cands_{\text{obs}}} s(y)$ 
\end{algorithmic}
}
\hrule
\vspace{1mm}

\captionof{algorithm}{The BayesOpt+GP+P algorithm. BayesOpt+GP is a special case of this where $\beta=0$.}
\label{alg:bayesopt}
\end{figure*}

Our main algorithm is an adaptation of BayesOpt with GPs as described in Section \ref{sec:bayesopt_gp} to the reranking setting. Each source sentence $x$ and its associated candidate list is treated as a standalone BayesOpt problem, meaning that no observations are shared across different $x$.
Thus for brevity, we omit $x$ from notation when discussing BayesOpt for a particular instance.

Let $s$ be the scoring function, an MT quality estimator. Let $\cands$ be a set of candidates, $\candsobs \subseteq \cands$ the subset of candidates for which we have observed $s(y)$, and $\candsobsbar$ be all other $y$ ($\candsobsbar = \cands \setminus \candsobs$). To perform reranking for an instance, we first generate candidates $\mathcal{C}$ and initialize the algorithm by scoring a random $\alpha$-sized subset of the list with $s$. In one iteration in the algorithm loop, we normalize the observed scores to mean 0 and 1 variance at every step and assume a 0 unconditional mean. Then we compute the GP posterior of all $y\in\candsobsbar$ with Equation \ref{eq:gaussian_posterior_mu} and \ref{eq:gaussian_posterior_sigma} given the scores of $\candsobs$, which is then used to compute EI with Equation \ref{eq:ei}, assuming no observation noise. We score the $k$ candidates in $\candsobsbar$ with the highest EI, adding them to $\candsobs$ (as well as removing them from $\candsobsbar$), and repeat the loop, terminating when a predefined budget of $n$ calls to $s$ is reached (or when all candidates have been evaluated, in the case that $|\mathcal{C}|\leq n$.). Finally, we choose $\arg\max_{y\in\candsobs}s(y)$ as the prediction.

We now describe our choice of GP kernel. $y\in\mathcal{C}$ are strings, and we seek a representation that is fast to compute and to compare, since $|\mathcal{C}|$ representations are generated, and the computing the GP covariance matrix requires $|\mathcal{C}|^2$ comparisons. Our kernel is $\mathcal{K}_\mathrm{MT}(y_i,y_j)=\mathcal{K}_\mathrm{RBF}(\mathrm{emb}(y_i),\mathrm{emb}(y_j))$, where $\mathrm{emb}$ returns the mean-pooled token-level outputs of the final decoder layer when generating $y$, normalized to the unit norm after pooling. $\mathrm{emb}$ uses meaning representations produced automatically during candidate list generation, so the additional cost to compute it is negligible. Also, the covariance matrix is fast to compute given the candidate list sizes and embedding dimensionality used in our experiments.

\subsection{Multi-fidelity BayesOpt}

We also propose an extension to BayesOpt+GP for the setting where observations are available from a different but related \textit{proxy score} function $s'$. We refer to this as BayesOpt+GP+P. $s'$ is assumed to have non-trivial covariance with the scoring model $s$ and to be cheaper to evaluate. This is known as \textit{multi-fidelity} BayesOpt in the literature, but while the multi-fidelity settings of \citet{NIPS2016_605ff764,pmlr-v115-wu20a} use acquisition functions that may choose to evaluate lower-fidelity scores, we study a simpler setting:  $\beta$ observations of $s'$ are obtained at the start where $\beta>\alpha$, and only $s$ may be evaluated during the BayesOpt loop. In the multi-fidelity setting, observations are made on $\langle y_i,s_i\rangle$, a combination of a data point and scoring function, instead of the data point alone.

Our kernel for BayesOpt+GP+P is the product of the RBF kernel from Section \ref{sec:bayesopt+gp} and a kernel over score functions $f$:
\begin{multline}
\hspace{-1mm}
\mathcal{K}_\mathrm{mult}\big(\langle y_i, s_k \rangle, \langle y_j, s_l \rangle\big)= \\
\mathcal{K}_\mathrm{MT}(y_i,y_j)\mathcal{K}_\mathrm{score}(s_k,s_l).
\hspace{-1mm}
\end{multline}
$\mathcal{K}_\mathrm{mult}$ is a valid kernel because a product of two kernels defined on different spaces is also a kernel \citep{rasmussen2005gaussian}.
With $\mathcal{K}_\mathrm{mult}$, the covariance between two observations depends on both the difference between scoring functions and the distance between data points. This way, an observation influences the posterior for all other data points at all choices of scoring function, as long as the scoring functions are correlated. This formulation enables the use of any number of scoring functions, but in this work, we consider at most two: the main scorer $s$ and a proxy scorer $s'$.

We set $\mathcal{K}_\mathrm{score}(s_k, s_l)$ to be the empirical covariance between $s_k$ and $s_l$ measured over a validation set, where all scores are normalized per-instance so that in each instance, the scores of all candidates for a particular scorer have 0 mean and 1 variance. Then for each scoring function, concatenate all candidate scores across instances, and compare the resulting lists to obtain the covariance. Covariance is a valid kernel because the covariance calculation can be expressed as a dot product, and dot products are valid kernels.

Proxy scores are incorporated into posterior estimation given by Equations \ref{eq:gaussian_posterior_mu} and \ref{eq:gaussian_posterior_sigma} by redefining $a$ to be a tuple of $\langle$data point, scoring function$\rangle$ and $\mathbf{a}$ to be a vector of such tuples. The kernel $\mathcal{K}$ is set to $\mathcal{K}_\mathrm{mult}$ which takes as input two tuples of data point and scoring function.
The full BayesOpt+GP+P algorithm is in Algorithm \ref{alg:bayesopt}.

\subsection{Proxy scores}
\label{sec:qe_models}
We train smaller scoring models to have high covariance with $s$ for use in BayesOpt+GP+P. In this work, our scoring functions are based on the Comet referenceless quality estimation architecture \citep{rei-etal-2020-comet}, also known as \cometkiwi.
These models encode the source and hypothesis jointly with a bidirectional transformer. 
Activations from all transformer layers are pooled to form a fixed-size representation, which is passed to a feed-forward regression head.
The vast majority of computation in this models is spent in the encoder.
Thus, faster Comet models can be obtained by reducing the size of the encoder.

We train Comet models using two differently sized pretrained multilingual encoder models in two ways: (1) training on the same training set as \cometkiwi and (2) distillation.
Among distillation methods, we attempt in preliminary experiments (1) training on the same training set as \cometkiwi with ground truth scores replaced with \cometkiwi scores and (2) training on a synthetic dataset comprising of LM samples along with their associated \cometkiwi scores.
The latter achieves higher correlation with \cometkiwi on sampled candidates, which is to be expected since the training distribution is more suitable for the reranking use case.
We therefore use this latter distillation method for all subsequent experiments. A similar procedure has been described in \citet{rei-etal-2022-searching}.

\subsection{Candidate list generation}\label{subsec:cand_list}

In preliminary experiments, we consider generating the candidate list using beam search with 128 outputs versus sampling 200 candidates using $\epsilon$-sampling \citep{hewitt-etal-2022-truncation} with $\epsilon=0.02$, a setting which effectively balances quality and diversity for MBR \citep{freitag-etal-2023-epsilon}. Under beam search, the candidates exhibit high lexical overlap, and while the mean score of candidates is higher, the average maximum score is lower.
The effectiveness of truncated sampling over beam search in larger conditional language model has also been observed by \citet{fernandes-etal-2022-quality}.

Furthermore, beam search suffers from out-of-memory errors on long translations, whereas with sampling, we simply reduce the batch size when out of memory. While it is possible to implement beam search in a batched manner, this does not exist in any popular conditional language model libraries, to the best of our knowledge.

For these reasons, we generate 200 candidates per instance with $\epsilon$-sampling, $\epsilon=0.02$ in all experiments. The sampled candidate list is then deduplicated, resulting in $\sim$178 candidates on average per instance.

\begin{figure*}
    \centering
    \includegraphics[width=1\linewidth]{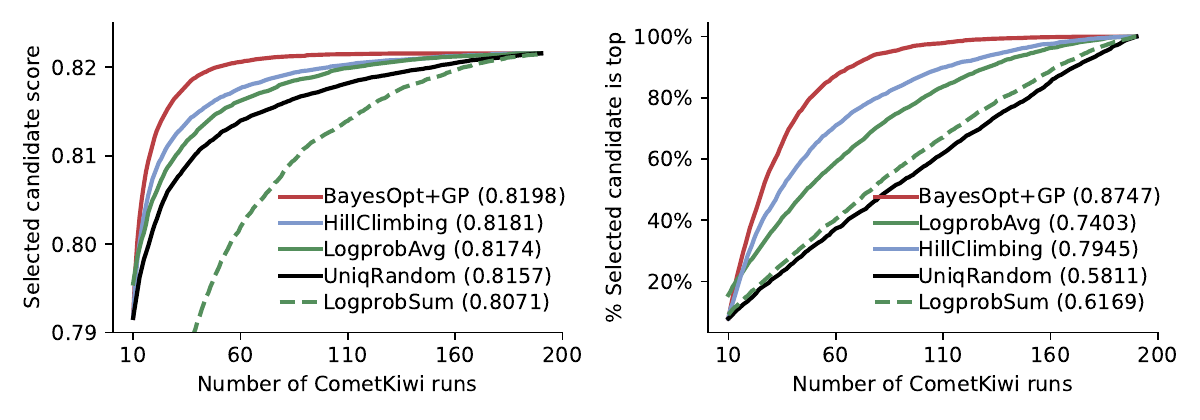}
    
    \caption{
    Left: Performance of reranking methods measured as the average \cometkiwi score of the selected candidate. Beam search with beam size 5 achieves a score of 0.754 and is too low to be pictured here.
    Right: percentage of instances where the selected candidate had the highest score (right).
    The x-axis is the scoring budget.
    Legends show the normalized area under the curve of \cometkiwi score of each method in brackets.  
    }
    \label{fig:results_baselines}
    
\end{figure*}

\section{Experiments}

We now discuss the details and findings of our Bayesian optimization experiments, followed by analysis of our trained proxy scoring models, concluding with runtime measurements. All run time values are measured on a A100-SXM4-40GB GPU.
For exact values for figures in this section, see \Cref{tab:results_exact_values} in the Appendix.
\Cref{sec:statistical_significance} contains extensive statistical significance tests.

For BayesOpt experiments, we grid search for the optimal value of RBF bandwidth parameter $w$ on the entire validation set, setting scoring budget $n=100$ and batch size $k=1$.
While it is possible to optimize it for every unique combination of language pair, $n$, $k$, proxy scoring function, and $\beta$, we find that the results are not statistically significantly different within a range of settings.
For simplicity, and to demonstrate the robustness of our methods, we use the same $w$ for all experiments. 

In all experiments, we use $\alpha=10$ initial randomly scored candidates. We set $k=1$ in \Cref{sec:experiments_bayesopt,sec:experiments_baseopt_proxy} to demonstrate the effectiveness of BayesOpt+GP under ideal conditions, but since $k$ can have a large impact on speed, we experiment with varying it in \Cref{sec:batch_size}.

\subsection{Models and datasets}\label{subsec:models_dataset}

For candidate generation, we use the \href{https://huggingface.co/facebook/nllb-200-distilled-600M}{600M-parameter distilled NLLB model} \citep{nllb2022} in all experiments.
For the main scoring model, we use \href{https://huggingface.co/Unbabel/wmt22-cometkiwi-da}{\cometkiwi-22} \citep{rei-etal-2022-cometkiwi}.

As a dataset used for proxy model training, we use data from the WMT Metrics Shared Task up to 2022 \citep{freitag-etal-2023-results}, which contains tuples of $\langle$source, hypothesis, human score$\rangle$.
The human scores were largely collected with the DA+SQM annotation protocol \citep{kocmi-etal-2022-findings}.

For BayesOpt experiments, we select the first 1000 and 500 source sentences per language pair from the WMT23 Metrics Shared Task dataset as the validation and test set, respectively, for 7 language pairs: English-Czech, English-German, English-Japanese, English-Chinese, and the reverse directions of the latter 3 pairs.

\cometkiwi is based on the encoder of \href{https://huggingface.co/FacebookAI/xlm-roberta-large}{XLM-Roberta\textsubscript{large}} \citep{DBLP:journals/corr/abs-1911-02116} (2.2GB memory).
For proxy scorers we train smaller models based on \href{https://huggingface.co/FacebookAI/xlm-roberta-large}{XLM-Roberta\textsubscript{base}} (1.1GB), and \href{https://huggingface.co/microsoft/Multilingual-MiniLM-L12-H384}{Multilingual-MiniLM-L12-H384} \citep{wang2020minilm} (469MB).

\subsection{BayesOpt+GP}
\label{sec:experiments_bayesopt}

\begin{figure*}[t]
    \centering
    \includegraphics[width=0.49\linewidth]{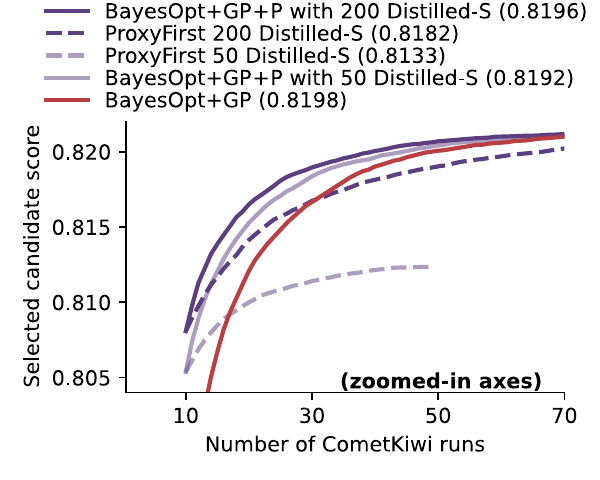}
    \hfill
    \includegraphics[width=0.49\linewidth]{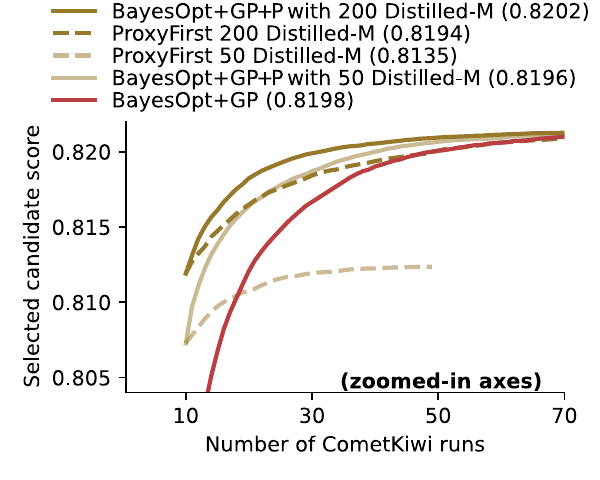}
    
    \caption{Average \cometkiwi score of the selected top candidate (y-axis) for BayesOpt+GP+P with Distilled-S (left) and Distilled-M (right) compared to the ProxyFirst baseline. This figure disregards the additional compute costs for these proxy metrics in order to show the marginal score increase from proxy observations.
    }
    \label{fig:results_multi_SM}
\end{figure*}

\begin{figure}[h]
    \centering
    \includegraphics[width=1\linewidth]{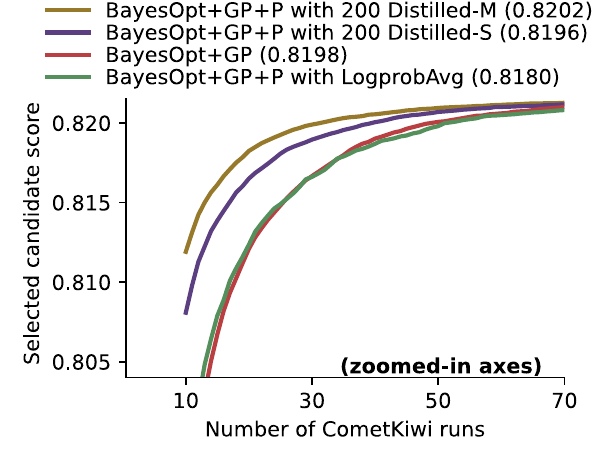}
    
    \caption{Average \cometkiwi score of the selected candidate (y-axis) for BayesOpt+GP+P with different choices of proxy score.}
    \label{fig:results_multi}
    
\end{figure}

The goal of reranking BayesOpt+GP is to improve the speed by only evaluating a subset of available candidates.
We evaluate this through quality-cost tradeoff curves, where quality is determined by final selected candidate's \cometkiwi score, and cost is determined by the number of calls to the scoring function.
As another measure of approximation quality, we also show the percentage of instances in which the actual best scoring candidate is returned. We devise several baselines with which to compare BayesOpt+GP. Each is a strategy for selecting a subset of candidates to score from which the best scoring candidate is returned. The baselines are:

\begin{itemize}
    \item \textbf{UniqRandom}: Shuffle the candidate list \textit{before} de-duplication, then de-duplicate while preserving the order of the first appearance of each candidate. Select the first $\min(n,|\mathcal{C}|)$ candidates in the resulting list.
    \item \textbf{Logprob\{Avg,Sum\}}: Sort $\mathcal{C}$ in order of negative sequence log probability (either average or sum), and then select the first $\min(n,|\mathcal{C}|)$.
    \item \textbf{HillClimbing}: Let $y^+$ be the highest scoring observation point at any time step. Iteratively select $\arg \min_{y\in\candsobsbar}||\mathrm{emb}(y)-\mathrm{emb}(y^+)||$ as the next observation point until $\min(n,|\mathcal{C}|)$ candidates are scored.
\end{itemize}
UniqRandom simulates the effect of iteratively sampling candidates until $n$ unique candidates are obtained. LogprobFirst\{Avg,Sum\} are included to verify whether more advanced methods indeed outperform simple subset selection using statistics obtained for free. HillClimbing is a heuristic iterative selection strategy which, like BayesOpt, is black-box and derivative-free \citep{conn2009derivative}.

In \Cref{fig:results_baselines}, BayesOpt+GP outperforms all baselines, and HillClimbing is the best among the baselines, with LogprobAvg following behind. LogprobSum severely underperforms UniqRandom in score, confirming findings on the inadequacy of very high probability translations \citep{eikema-aziz-2020-map}. Informally speaking, UniqRandom is a simple ``exploration'' strategy that ignores existing observations, while HillClimbing is a simple ``exploitation'' strategy, only searching over neighbors nearest the best observation while ignoring the full search space. These results confirm that balancing these respective deciderata helps to find the optimal candidate more efficiently.

\subsection{BayesOpt+GP+P}
\label{sec:experiments_baseopt_proxy}

\subsubsection{Proxy score evaluation}

We first evaluate trained proxy scorers independently of their use in BayesOpt according to
(1) actual runtime,
(2) correlation with human ratings in the WMT23 dataset,
(3) correlation with \cometkiwi on source-hypothesis pairs in WMT23, and
(4) correlation with \cometkiwi on a synthetic candidates for an instance, averaged over instances. For correlations we use Kendall's $\tau_c$, which is commonly used in MT metric evaluation \citep{freitag-etal-2023-results}.

\Cref{tab:qe_models_main} shows the results for the proxy models.
The model size corresponds closely to inference time.
As desired, training proxies using distillation results in much higher correlation with \cometkiwi, although it loses some correlation with human judgments. In subsequent experiments, we consider Distilled-\{S,M\} only. While LogprobAvg has comparatively much lower correlation, we nevertheless consider it as a proxy score since it is obtained for free during candidate generation.

\begin{table}[htbp]
\centering
\resizebox{\linewidth}{!}{
\begin{tabular}{lccccc}
\toprule
&& \bf Human & \multicolumn{2}{c}{\bf \cometkiwi} \\
\bf Model & \bf Time & \bf Test & \bf Test & \bf Cands. \\
\midrule
\cometkiwi & 51.38s & 0.245 & 1.000 & 1.000 \\
\cmidrule{1-1}
LogprobsAvg & \hphantom{0}0.00s & - & - & 0.191 \\
LogprobsSum & \hphantom{0}0.00s & - & - &\hspace{-1.1mm}-0.090 \\
\cmidrule{1-1}
Authentic-S & \hphantom{0}7.13s & 0.193 & 0.314 & 0.350 \\
Authentic-M & 18.71s & 0.199 & 0.320 & 0.448 \\
\cmidrule{1-1}
Distilled-S & \hphantom{0}7.13s & 0.169 & 0.488 & 0.620 \\
Distilled-M & 18.71s & 0.188 & 0.572 & 0.680 \\
\bottomrule
\end{tabular}
}

\caption{
Benchmarking proxy models (\Cref{sec:qe_models}) on speed and correlation with human judgments/\cometkiwi using the WMT23 dataset. Speed is measured by runtime per 10000 samples using maximum batch size.
Correlation is measured with Kendall's $\tau_c$ against human judgments and \cometkiwi scores. \cometkiwi correlation is taken over the provided targets in WMT23 (Test) and a synthetic dataset comprised of 200 samples per source sentence, deduped (Cands).
Logprobs\{Avg,Sum\} is not evaluated on WMT23 targets because they are generated by other MT systems.
}
\label{tab:qe_models_main}
\end{table}

\subsubsection{Reranking results}

When $s'$ is sufficiently fast and correlated with $s$, it can further improve the quality-cost tradeoff in BayesOpt+GP.
Recall that BayesOpt+GP+P initializes with $\beta$ evaluations of $s'$.
\Cref{fig:results_multi} shows the quality-cost curve when all proxy scores are known, or $\beta=200$.
The relative performance when including proxy scores correspond to their correlation with \cometkiwi as shown in \Cref{tab:qe_models_main}; Distilled-M outperforms Distilled-S, and both outperform LogprobAvg.
This demonstrates the importance of ensuring high correlation in the proxy score.
The addition of LogprobAvg to BayesOpt+GP has little effect, showing that poorly correlated proxies are too noisy to help and may even hinder performance.
Beyond $n=70$, all methods achieve close to the maximum attainable score.

We also examine the effect of initializing with a fraction of proxy observations rather than all of them.
For some choice of $\beta$, an appropriate baseline is to rank the top-$n$ candidates among the $\beta$ observed proxy scores.
We call this ProxyFirst.
The results when using Distilled-M and Distilled-S as proxies are shown in \Cref{fig:results_multi_SM}.
In both cases, the difference between BayesOpt+GP+P and ProxyFirst is smaller when $\beta=200$ than when $\beta=50$, and this gap is smaller for Distilled-M.
This is to be expected because as the covariance of $s$ and $s'$ increases, using ProxyFirst with $\beta=200$ approaches standard full-list reranking.
The marginal benefit of BayesOpt+GP+P is more clear when $\beta=50$, where proxy scores help to find promising candidates earlier.

Overall, proxy observations can indeed improve quality for a particular $n$.
However, for sufficiently large $n$, BayesOpt+GP converges, so proxy observations are unnecessary.
Proxy evaluations add to the runtime cost which we discuss in \Cref{sec:runtime}.
Therefore, while we show that the multi-fidelity kernel is capable of leveraging proxy scores to improve search, in practice, the overall computational budget should be considered along with the quality and cost of the proxy scoring function to ensure that using the method is worthwhile.

\subsection{Runtime}
\label{sec:runtime}

Our reranking algorithm significantly reduces actual runtime compared to scoring all candidates for a source sentence. We profile the full pipeline, from generating candidates to making a final selection, on three settings: (1) BayesOpt+GP with $n=90$, and (2) multi-fidelity BayesOpt+GP with 50 Distilled-S scores and $n=70$, and 3) the baseline of evaluating \cometkiwi on all candidates. $n,\beta$ are selected to balance the final scores of the two algorithms ($0.8213$ and $0.8211$ respectively, as shown in Table \ref{tab:results_exact_values}).

For the runtime calculations, we select 50 source sentences from each language pair and generate 200 candidates for each.
For the baseline, we compute scores for all candidates with a batch size of 200.
For BayesOpt+GP methods, we profile the additional steps required: computing the kernel, computing the posteriors at each step, and evaluating proxy scores. BayesOpt+GP(+S) uses batch size $k=10$, which does not affect scores compared to using $k=1$ (see  \Cref{sec:batch_size}). Memory bandwidth can be a major overhead in large neural networks, making it inefficient to run small batches. Since BayesOpt+GP obtains $k$ candidates per step, in order to use large batches, we process candidates for multiple instances in parallel.

Results are shown in Table \ref{tab:runtimes}. In all cases, candidate generation and \cometkiwi calculations dominate the overall runtime. 
The extra cost from BayesOpt-related computations is compensated by the savings from reducing \cometkiwi evaluations, despite similarity matrix computation being $\mathcal{O}(|\mathcal{C}|^2)$, and matrix inversion for posterior calculation at each iteration being $\mathcal{O}(|\mathcal{C}|^3)$.
BayesOpt+GP+P with Distilled-S reduces the runtime by further reducing the number of \cometkiwi calculations to 70, with the cost of loading and running the Distilled-S proxy metric introducing minimal overhead.

\begin{table}[htbp]
\centering
\resizebox{\linewidth}{!}{
\renewcommand{\arraystretch}{1.1}
\begin{tabular}{l>{\hspace{-2mm}}cccc}
\toprule
\bf   & \bf  & \bf BayesOpt & \bf BayesOpt \\
\bf Operation & \bf AllComet & \bf +GP & \bf +GP+P \\
\bf  & \bf  & \bf $n=90$ & \bf $n=70,\beta=50$ \\
\midrule
Candidates  & 701.38 & 701.38  & 701.38 \\
Similarities & - & \hphantom{00}1.24 & \hphantom{00}1.24  \\
BayesOpt+GP  & - & \hphantom{00}1.92 & \hphantom{00}2.25 \\
Comet Loading & \hphantom{0}8.43 & \hphantom{00}8.43 & \hphantom{0}11.27 \\
Distilled-S & - & - & \hphantom{0}11.11  \\
\cometkiwi & 274.87 & 188.39 & 146.33 \\
\cmidrule{1-1}
Total & 984.68 &  901.36 & 873.58\\
\bottomrule
\end{tabular}
}

\caption{Runtimes for the full reranking baseline (AllComet), BayesOpt+GP, and BayesOpt+GP+P with Distilled-S as proxy score at settings where \cometkiwi scores are roughly equal. Time given in seconds per 350 instances.}
\label{tab:runtimes}
\end{table}

\subsection{Batch size $\bm{k}$ in BayesOpt+GP}
\label{sec:batch_size}

\begin{figure}[h]
    \centering
    \includegraphics[width=0.95\linewidth]{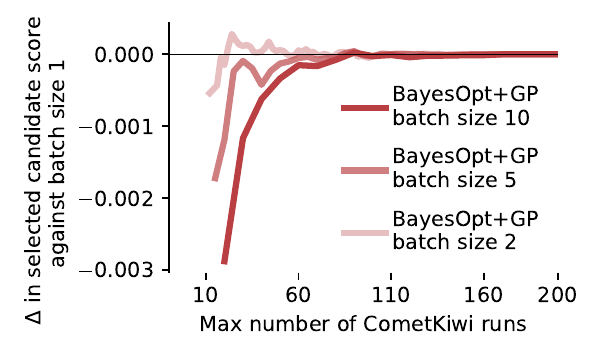}    
    \caption{Difference between BayesOpt+GP with batch size of 1 (top line in red in \Cref{fig:results_baselines}) and BayesOpt+GP with higher batch sizes. Negative values mean that higher batch size performed worse than BayesOpt+GP with batch size of 1.}
    \label{fig:batch_size}

\end{figure}

We examine the effect of batch size $k$ in BayesOpt+GP for $k=1,2,5,10$.
Figure \ref{fig:batch_size} shows that as expected, larger $k$ diminishes performance, although the differences nearly vanish at $n{>}70$. 

$k$ impacts how often the BayesOpt loop is run and thus has a large effect on speed. Fortunately, we observe for sufficiently large $n$, $k$ can be increased without sacrificing quality.

\section{Conclusion}

In this work, we formalize MT reranking as a Bayesian optimization problem, leveraging the basic observation that similar translations are more likely to have similar quality scores. We also extend the framework to accept observations from proxy scoring functions, which is applicable when the target score is very costly: large QE models, MBR, or human evaluation. In realistic experiments, we show that our methods improve reranking efficiency over strong baselines. We also propose several design choices that make the methods useful in practice; a GP kernel that requires minimal overhead, and effective proxy model training via distillation.

We consider our work a first step in applying BayesOpt to MT reranking. Future directions include integrating BayesOpt with candidate generation, alternative acquisition functions, and further exploration of GP kernels for MT.

\section{Limitations}

The optimization problem considered in this work is to maximize score from a scoring model. We show that BayesOpt is an effective optimizer, but we do not explore to what extent the optimization problem is flawed due to flaws in the scoring model. We refer to \citet{kocmi-etal-2024-navigating} to understand what magnitude of score difference between systems is significant. However, the existence of ``metric overfitting'' when directly optimizing an evaluation metric is debated and may affect the interpretation of score differences \citep{fernandes-etal-2022-quality,wang2024dontthrowawaydata}.

BayesOpt+GP requires matrix inversion, a $\mathcal{O}(|\mathcal{C}|^3)$ operation that is performed once per iteration. While it is inexpensive for the $|\mathcal{C}|$ we consider, this limits the number of observations that can be used for posterior computation without resorting to approximations \citep{osti_1963337}.

As an iterative algorithm, BayesOpt can score no more than $k$ candidates in a batch for a single instance. Small batch sizes introduce a significant bottleneck for large neural networks, so in order to maintain large batch sizes, we propose processing multiple instances in parallel. However, this requires additional engineering.

\section*{Acknowledgements}

Julius Cheng is supported by a scholarship from Huawei. Part of this work received support from the European Union’s Horizon research and
innovation programme under grant agreement No 101135798, project Meetween (My Personal AI Mediator for Virtual MEETtings BetWEEN People).
We thank the organizers of MT Marathon 2024, where the authors met and this work was conceived.
We also thank Béni Egressy for useful discussions and Will Tebbutt for lending expertise on GPs.

\bibliography{misc/anthology.min.bib,misc/bibliography.bib}

\begin{thebibliography}{52}
\expandafter\ifx\csname natexlab\endcsname\relax\def\natexlab#1{#1}\fi

\bibitem[{Beck and Cohn(2017)}]{beck-cohn-2017-learning}
Daniel Beck and Trevor Cohn. 2017.
\newblock \href {https://aclanthology.org/I17-2012} {Learning kernels over strings using {G}aussian processes}.
\newblock In \emph{Proceedings of the Eighth International Joint Conference on Natural Language Processing (Volume 2: Short Papers)},  67--73. Asian Federation of Natural Language Processing.

\bibitem[{Beck et~al.(2014)Beck, Cohn, and Specia}]{beck-etal-2014-joint}
Daniel Beck, Trevor Cohn, and Lucia Specia. 2014.
\newblock \href {https://doi.org/10.3115/v1/D14-1190} {Joint emotion analysis via multi-task {G}aussian processes}.
\newblock In \emph{Proceedings of the 2014 Conference on Empirical Methods in Natural Language Processing ({EMNLP})},  1798--1803. Association for Computational Linguistics.

\bibitem[{Beck et~al.(2013)Beck, Shah, Cohn, and Specia}]{beck-etal-2013-shef}
Daniel Beck, Kashif Shah, Trevor Cohn, and Lucia Specia. 2013.
\newblock \href {https://aclanthology.org/W13-2241} {{SHEF}-{L}ite: When less is more for translation quality estimation}.
\newblock In \emph{Proceedings of the Eighth Workshop on Statistical Machine Translation},  337--342. Association for Computational Linguistics.

\bibitem[{Bergstra et~al.(2011)Bergstra, Bardenet, Bengio, and K\'{e}gl}]{bergstra2011hyperopt}
James Bergstra, R\'{e}mi Bardenet, Yoshua Bengio, and Bal\'{a}zs K\'{e}gl. 2011.
\newblock \href {https://proceedings.neurips.cc/paper_files/paper/2011/file/86e8f7ab32cfd12577bc2619bc635690-Paper.pdf} {Algorithms for hyper-parameter optimization}.
\newblock In \emph{Advances in Neural Information Processing Systems}, volume~24. Curran Associates, Inc.

\bibitem[{Bhattacharyya et~al.(2021)Bhattacharyya, Rooshenas, Naskar, Sun, Iyyer, and McCallum}]{bhattacharyya-etal-2021-energy}
Sumanta Bhattacharyya, Amirmohammad Rooshenas, Subhajit Naskar, Simeng Sun, Mohit Iyyer, and Andrew McCallum. 2021.
\newblock \href {https://doi.org/10.18653/v1/2021.acl-long.349} {Energy-based reranking: Improving neural machine translation using energy-based models}.
\newblock In \emph{Proceedings of the 59th Annual Meeting of the Association for Computational Linguistics and the 11th International Joint Conference on Natural Language Processing (Volume 1: Long Papers)},  4528--4537. Association for Computational Linguistics.

\bibitem[{Charniak and Johnson(2005)}]{charniak-johnson-2005-coarse}
Eugene Charniak and Mark Johnson. 2005.
\newblock \href {https://doi.org/10.3115/1219840.1219862} {Coarse-to-fine n-best parsing and {M}ax{E}nt discriminative reranking}.
\newblock In \emph{Proceedings of the 43rd Annual Meeting of the Association for Computational Linguistics ({ACL}{'}05)},  173--180. Association for Computational Linguistics.

\bibitem[{Chen et~al.(2023)Chen, Zaharia, and Zou}]{chen2023frugalgptuselargelanguage}
Lingjiao Chen, Matei Zaharia, and James Zou. 2023.
\newblock \href {http://arxiv.org/abs/2305.05176} {Frugalgpt: How to use large language models while reducing cost and improving performance}.

\bibitem[{Cheng and Vlachos(2023)}]{cheng-vlachos-2023-faster}
Julius Cheng and Andreas Vlachos. 2023.
\newblock \href {https://doi.org/10.18653/v1/2023.emnlp-main.767} {Faster minimum {B}ayes risk decoding with confidence-based pruning}.
\newblock In \emph{Proceedings of the 2023 Conference on Empirical Methods in Natural Language Processing},  12473--12480. Association for Computational Linguistics.

\bibitem[{Collins and Koo(2005)}]{collins-koo-2005-discriminative}
Michael Collins and Terry Koo. 2005.
\newblock \href {https://doi.org/10.1162/0891201053630273} {Discriminative reranking for natural language parsing}.
\newblock \emph{Computational Linguistics}, 31(1):25--70.

\bibitem[{Conn et~al.(2009)Conn, Scheinberg, and Vicente}]{conn2009derivative}
Andrew~R. Conn, Katya Scheinberg, and Luis~N. Vicente. 2009.
\newblock \href {https://doi.org/10.1137/1.9780898718768} {\emph{Introduction to Derivative-Free Optimization}}.
\newblock Society for Industrial and Applied Mathematics.

\bibitem[{Conneau et~al.(2019)Conneau, Khandelwal, Goyal, Chaudhary, Wenzek, Guzm{\'{a}}n, Grave, Ott, Zettlemoyer, and Stoyanov}]{DBLP:journals/corr/abs-1911-02116}
Alexis Conneau, Kartikay Khandelwal, Naman Goyal, Vishrav Chaudhary, Guillaume Wenzek, Francisco Guzm{\'{a}}n, Edouard Grave, Myle Ott, Luke Zettlemoyer, and Veselin Stoyanov. 2019.
\newblock \href {http://arxiv.org/abs/1911.02116} {Unsupervised cross-lingual representation learning at scale}.
\newblock \emph{CoRR}, abs/1911.02116.

\bibitem[{Deguchi et~al.(2024)Deguchi, Sakai, Kamigaito, Watanabe, Tanaka, and Utiyama}]{deguchi-etal-2024-centroid}
Hiroyuki Deguchi, Yusuke Sakai, Hidetaka Kamigaito, Taro Watanabe, Hideki Tanaka, and Masao Utiyama. 2024.
\newblock \href {https://doi.org/10.18653/v1/2024.findings-acl.654} {Centroid-based efficient minimum {B}ayes risk decoding}.
\newblock In \emph{Findings of the Association for Computational Linguistics ACL 2024},  11009--11018, Bangkok, Thailand and virtual meeting. Association for Computational Linguistics.

\bibitem[{Eikema and Aziz(2020)}]{eikema-aziz-2020-map}
Bryan Eikema and Wilker Aziz. 2020.
\newblock \href {https://doi.org/10.18653/v1/2020.coling-main.398} {Is {MAP} decoding all you need? the inadequacy of the mode in neural machine translation}.
\newblock In \emph{Proceedings of the 28th International Conference on Computational Linguistics},  4506--4520. International Committee on Computational Linguistics.

\bibitem[{Eikema and Aziz(2022)}]{eikema-aziz-2022-sampling}
Bryan Eikema and Wilker Aziz. 2022.
\newblock \href {https://doi.org/10.18653/v1/2022.emnlp-main.754} {Sampling-based approximations to minimum {B}ayes risk decoding for neural machine translation}.
\newblock In \emph{Proceedings of the 2022 Conference on Empirical Methods in Natural Language Processing},  10978--10993. Association for Computational Linguistics.

\bibitem[{Fernandes et~al.(2022)Fernandes, Farinhas, Rei, C.~de Souza, Ogayo, Neubig, and Martins}]{fernandes-etal-2022-quality}
Patrick Fernandes, Ant{\'o}nio Farinhas, Ricardo Rei, Jos{\'e}~G. C.~de Souza, Perez Ogayo, Graham Neubig, and Andre Martins. 2022.
\newblock \href {https://doi.org/10.18653/v1/2022.naacl-main.100} {Quality-aware decoding for neural machine translation}.
\newblock In \emph{Proceedings of the 2022 Conference of the North American Chapter of the Association for Computational Linguistics: Human Language Technologies},  1396--1412. Association for Computational Linguistics.

\bibitem[{Finkelstein et~al.(2024)Finkelstein, Naskar, Mirzazadeh, Shah, and Freitag}]{finkelstein2024mbrqefinetuningtrainingtime}
Mara Finkelstein, Subhajit Naskar, Mehdi Mirzazadeh, Apurva Shah, and Markus Freitag. 2024.
\newblock \href {http://arxiv.org/abs/2309.10966} {{MBR} and {QE} finetuning: {Training-time} distillation of the best and most expensive decoding methods}.

\bibitem[{Freitag et~al.(2023{\natexlab{a}})Freitag, Ghorbani, and Fernandes}]{freitag-etal-2023-epsilon}
Markus Freitag, Behrooz Ghorbani, and Patrick Fernandes. 2023{\natexlab{a}}.
\newblock \href {https://doi.org/10.18653/v1/2023.findings-emnlp.617} {Epsilon sampling rocks: Investigating sampling strategies for minimum {B}ayes risk decoding for machine translation}.
\newblock In \emph{Findings of the Association for Computational Linguistics: EMNLP 2023},  9198--9209. Association for Computational Linguistics.

\bibitem[{Freitag et~al.(2022)Freitag, Grangier, Tan, and Liang}]{freitag-etal-2022-high}
Markus Freitag, David Grangier, Qijun Tan, and Bowen Liang. 2022.
\newblock \href {https://doi.org/10.1162/tacl\_a\_00491} {High quality rather than high model probability: Minimum {B}ayes risk decoding with neural metrics}.
\newblock \emph{Transactions of the Association for Computational Linguistics}, 10:811--825.

\bibitem[{Freitag et~al.(2023{\natexlab{b}})Freitag, Mathur, Lo, Avramidis, Rei, Thompson, Kocmi, Blain, Deutsch, Stewart, Zerva, Castilho, Lavie, and Foster}]{freitag-etal-2023-results}
Markus Freitag, Nitika Mathur, Chi-kiu Lo, Eleftherios Avramidis, Ricardo Rei, Brian Thompson, Tom Kocmi, Frederic Blain, Daniel Deutsch, Craig Stewart, Chrysoula Zerva, Sheila Castilho, Alon Lavie, and George Foster. 2023{\natexlab{b}}.
\newblock \href {https://doi.org/10.18653/v1/2023.wmt-1.51} {Results of {WMT}23 metrics shared task: Metrics might be guilty but references are not innocent}.
\newblock In \emph{Proceedings of the Eighth Conference on Machine Translation},  578--628. Association for Computational Linguistics.

\bibitem[{Hennig and Schuler(2011)}]{Hennig2011EntropySF}
Philipp Hennig and Christian~J. Schuler. 2011.
\newblock \href {https://api.semanticscholar.org/CorpusID:166832} {Entropy search for information-efficient global optimization}.
\newblock \emph{ArXiv}, abs/1112.1217.

\bibitem[{Hewitt et~al.(2022)Hewitt, Manning, and Liang}]{hewitt-etal-2022-truncation}
John Hewitt, Christopher Manning, and Percy Liang. 2022.
\newblock \href {https://doi.org/10.18653/v1/2022.findings-emnlp.249} {Truncation sampling as language model desmoothing}.
\newblock In \emph{Findings of the Association for Computational Linguistics: EMNLP 2022},  3414--3427. Association for Computational Linguistics.

\bibitem[{Jones(2001)}]{Jones2001ATO}
Donald~R. Jones. 2001.
\newblock \href {https://api.semanticscholar.org/CorpusID:8723392} {A taxonomy of global optimization methods based on response surfaces}.
\newblock \emph{Journal of Global Optimization}, 21:345--383.

\bibitem[{Juraska et~al.(2023)Juraska, Finkelstein, Deutsch, Siddhant, Mirzazadeh, and Freitag}]{juraska-etal-2023-metricx}
Juraj Juraska, Mara Finkelstein, Daniel Deutsch, Aditya Siddhant, Mehdi Mirzazadeh, and Markus Freitag. 2023.
\newblock \href {https://doi.org/10.18653/v1/2023.wmt-1.63} {{M}etric{X}-23: The {G}oogle submission to the {WMT} 2023 metrics shared task}.
\newblock In \emph{Proceedings of the Eighth Conference on Machine Translation},  756--767. Association for Computational Linguistics.

\bibitem[{Kandasamy et~al.(2016)Kandasamy, Dasarathy, Oliva, Schneider, and Poczos}]{NIPS2016_605ff764}
Kirthevasan Kandasamy, Gautam Dasarathy, Junier~B Oliva, Jeff Schneider, and Barnabas Poczos. 2016.
\newblock \href {https://proceedings.neurips.cc/paper_files/paper/2016/file/605ff764c617d3cd28dbbdd72be8f9a2-Paper.pdf} {Gaussian process bandit optimisation with multi-fidelity evaluations}.
\newblock In \emph{Advances in Neural Information Processing Systems}, volume~29. Curran Associates, Inc.

\bibitem[{Kocmi et~al.(2024{\natexlab{a}})Kocmi, Avramidis, Bawden, Bojar, Dvorkovich, Federmann, Fishel, Freitag, Gowda, Grundkiewicz, Haddow, Karpinska, Koehn, Marie, Monz, Murray, Nagata, Popel, Popovi{\'c}, Shmatova, Steingr{\'i}msson, and Zouhar}]{kocmi-etal-2024-findings}
Tom Kocmi, Eleftherios Avramidis, Rachel Bawden, Ond{\v{r}}ej Bojar, Anton Dvorkovich, Christian Federmann, Mark Fishel, Markus Freitag, Thamme Gowda, Roman Grundkiewicz, Barry Haddow, Marzena Karpinska, Philipp Koehn, Benjamin Marie, Christof Monz, Kenton Murray, Masaaki Nagata, Martin Popel, Maja Popovi{\'c}, Mariya Shmatova, Steinth{\'o}r Steingr{\'i}msson, and Vil{\'e}m Zouhar. 2024{\natexlab{a}}.
\newblock \href {https://doi.org/10.18653/v1/2024.wmt-1.1} {Findings of the {WMT}24 general machine translation shared task: The {LLM} era is here but {MT} is not solved yet}.
\newblock In \emph{Proceedings of the Ninth Conference on Machine Translation},  1--46, Miami, Florida, USA. Association for Computational Linguistics.

\bibitem[{Kocmi et~al.(2022)Kocmi, Bawden, Bojar, Dvorkovich, Federmann, Fishel, Gowda, Graham, Grundkiewicz, Haddow, Knowles, Koehn, Monz, Morishita, Nagata, Nakazawa, Nov{\'a}k, Popel, and Popovi{\'c}}]{kocmi-etal-2022-findings}
Tom Kocmi, Rachel Bawden, Ond{\v{r}}ej Bojar, Anton Dvorkovich, Christian Federmann, Mark Fishel, Thamme Gowda, Yvette Graham, Roman Grundkiewicz, Barry Haddow, Rebecca Knowles, Philipp Koehn, Christof Monz, Makoto Morishita, Masaaki Nagata, Toshiaki Nakazawa, Michal Nov{\'a}k, Martin Popel, and Maja Popovi{\'c}. 2022.
\newblock \href {https://aclanthology.org/2022.wmt-1.1} {Findings of the 2022 conference on machine translation ({WMT}22)}.
\newblock In \emph{Proceedings of the Seventh Conference on Machine Translation (WMT)},  1--45. Association for Computational Linguistics.

\bibitem[{Kocmi et~al.(2024{\natexlab{b}})Kocmi, Zouhar, Federmann, and Post}]{kocmi-etal-2024-navigating}
Tom Kocmi, Vil{\'e}m Zouhar, Christian Federmann, and Matt Post. 2024{\natexlab{b}}.
\newblock \href {https://doi.org/10.18653/v1/2024.acl-long.110} {Navigating the metrics maze: {Reconciling} score magnitudes and accuracies}.
\newblock In \emph{Proceedings of the 62nd Annual Meeting of the Association for Computational Linguistics (Volume 1: Long Papers)},  1999--2014, Bangkok, Thailand. Association for Computational Linguistics.

\bibitem[{Krause et~al.(2008)Krause, Singh, and Guestrin}]{krause}
Andreas Krause, Ajit Singh, and Carlos Guestrin. 2008.
\newblock \href {https://www.jmlr.org/papers/volume9/krause08a/krause08a.pdf} {Near-optimal sensor placements in gaussian processes: {Theory,} efficient algorithms and empirical studies}.
\newblock \emph{J. Mach. Learn. Res.}, 9:235–284.

\bibitem[{Lee et~al.(2021)Lee, Auli, and Ranzato}]{lee-etal-2021-discriminative}
Ann Lee, Michael Auli, and Marc{'}Aurelio Ranzato. 2021.
\newblock \href {https://doi.org/10.18653/v1/2021.acl-long.563} {Discriminative reranking for neural machine translation}.
\newblock In \emph{Proceedings of the 59th Annual Meeting of the Association for Computational Linguistics and the 11th International Joint Conference on Natural Language Processing (Volume 1: Long Papers)},  7250--7264. Association for Computational Linguistics.

\bibitem[{Mockus(1974)}]{mockus1974bayesian}
Jonas Mockus. 1974.
\newblock \href {https://link.springer.com/content/pdf/10.1007/978-3-662-38527-2_55.pdf} {On bayesian methods for seeking the extremum}.
\newblock In \emph{Proceedings of the IFIP Technical Conference},  400--404.

\bibitem[{Noack et~al.(2023)Noack, Krishnan, Risser, and Reyes}]{osti_1963337}
Marcus~M. Noack, Harinarayan Krishnan, Mark~D. Risser, and Kristofer~G. Reyes. 2023.
\newblock \href {https://doi.org/10.1038/s41598-023-30062-8} {Exact gaussian processes for massive datasets via non-stationary sparsity-discovering kernels}.
\newblock \emph{Scientific Reports}, 13(1).

\bibitem[{Petrov(2011)}]{coarse-to-fine}
Slav Petrov. 2011.
\newblock \href {https://dl.acm.org/doi/abs/10.5555/2103611} {\emph{Coarse-to-Fine Natural Language Processing (Theory and Applications of Natural Language Processing)}}.
\newblock Springer Publishing Company, Incorporated.

\bibitem[{Rasmussen and Williams(2005)}]{rasmussen2005gaussian}
C.E. Rasmussen and C.K.I. Williams. 2005.
\newblock \href {https://books.google.co.uk/books?id=GhoSngEACAAJ} {\emph{Gaussian Processes for Machine Learning}}.
\newblock Adaptive Computation and Machine Learning series. MIT Press.

\bibitem[{Ravaut et~al.(2022)Ravaut, Joty, and Chen}]{ravaut-etal-2022-summareranker}
Mathieu Ravaut, Shafiq Joty, and Nancy Chen. 2022.
\newblock \href {https://doi.org/10.18653/v1/2022.acl-long.309} {{S}umma{R}eranker: A multi-task mixture-of-experts re-ranking framework for abstractive summarization}.
\newblock In \emph{Proceedings of the 60th Annual Meeting of the Association for Computational Linguistics (Volume 1: Long Papers)},  4504--4524. Association for Computational Linguistics.

\bibitem[{Rei et~al.(2022{\natexlab{a}})Rei, Farinha, de~Souza, Ramos, Martins, Coheur, and Lavie}]{rei-etal-2022-searching}
Ricardo Rei, Ana~C Farinha, Jos{\'e}~G.C. de~Souza, Pedro~G. Ramos, Andr{\'e}~F.T. Martins, Luisa Coheur, and Alon Lavie. 2022{\natexlab{a}}.
\newblock \href {https://aclanthology.org/2022.eamt-1.9} {Searching for {COMETINHO}: The little metric that could}.
\newblock In \emph{Proceedings of the 23rd Annual Conference of the European Association for Machine Translation},  61--70. European Association for Machine Translation.

\bibitem[{Rei et~al.(2024)Rei, Pombal, Guerreiro, Alves, Martins, Fernandes, Wu, Vaz, Alves, Farajian, Agrawal, Farinhas, C.~De~Souza, and Martins}]{rei-etal-2024-tower}
Ricardo Rei, Jose Pombal, Nuno~M. Guerreiro, Jo{\~a}o Alves, Pedro~Henrique Martins, Patrick Fernandes, Helena Wu, Tania Vaz, Duarte Alves, Amin Farajian, Sweta Agrawal, Antonio Farinhas, Jos{\'e}~G. C.~De~Souza, and Andr{\'e} Martins. 2024.
\newblock \href {https://doi.org/10.18653/v1/2024.wmt-1.12} {Tower v2: Unbabel-{IST} 2024 submission for the general {MT} shared task}.
\newblock In \emph{Proceedings of the Ninth Conference on Machine Translation},  185--204, Miami, Florida, USA. Association for Computational Linguistics.

\bibitem[{Rei et~al.(2020)Rei, Stewart, Farinha, and Lavie}]{rei-etal-2020-comet}
Ricardo Rei, Craig Stewart, Ana~C Farinha, and Alon Lavie. 2020.
\newblock \href {https://doi.org/10.18653/v1/2020.emnlp-main.213} {{COMET}: A neural framework for {MT} evaluation}.
\newblock In \emph{Proceedings of the 2020 Conference on Empirical Methods in Natural Language Processing (EMNLP)},  2685--2702. Association for Computational Linguistics.

\bibitem[{Rei et~al.(2022{\natexlab{b}})Rei, Treviso, Guerreiro, Zerva, Farinha, Maroti, C.~de Souza, Glushkova, Alves, Coheur, Lavie, and Martins}]{rei-etal-2022-cometkiwi}
Ricardo Rei, Marcos Treviso, Nuno~M. Guerreiro, Chrysoula Zerva, Ana~C Farinha, Christine Maroti, Jos{\'e}~G. C.~de Souza, Taisiya Glushkova, Duarte Alves, Luisa Coheur, Alon Lavie, and Andr{\'e} F.~T. Martins. 2022{\natexlab{b}}.
\newblock \href {https://aclanthology.org/2022.wmt-1.60} {{C}omet{K}iwi: {IST}-unbabel 2022 submission for the quality estimation shared task}.
\newblock In \emph{Proceedings of the Seventh Conference on Machine Translation (WMT)},  634--645. Association for Computational Linguistics.

\bibitem[{Sellam et~al.(2020)Sellam, Das, and Parikh}]{sellam-etal-2020-bleurt}
Thibault Sellam, Dipanjan Das, and Ankur Parikh. 2020.
\newblock \href {https://doi.org/10.18653/v1/2020.acl-main.704} {{BLEURT}: Learning robust metrics for text generation}.
\newblock In \emph{Proceedings of the 58th Annual Meeting of the Association for Computational Linguistics},  7881--7892. Association for Computational Linguistics.

\bibitem[{Shahriari et~al.(2016)Shahriari, Swersky, Wang, Adams, and de~Freitas}]{Shahriari2016TakingTH}
Bobak Shahriari, Kevin Swersky, Ziyun Wang, Ryan~P. Adams, and Nando de~Freitas. 2016.
\newblock \href {https://api.semanticscholar.org/CorpusID:14843594} {Taking the human out of the loop: {A} review of bayesian optimization}.
\newblock \emph{Proceedings of the IEEE}, 104:148--175.

\bibitem[{Singhal et~al.(2023)Singhal, Xu, Ye, and Durrett}]{singhal-etal-2023-eel}
Prasann Singhal, Jiacheng Xu, Xi~Ye, and Greg Durrett. 2023.
\newblock \href {https://doi.org/10.18653/v1/2023.acl-long.517} {{EEL}: Efficiently encoding lattices for reranking}.
\newblock In \emph{Proceedings of the 61st Annual Meeting of the Association for Computational Linguistics (Volume 1: Long Papers)},  9299--9316. Association for Computational Linguistics.

\bibitem[{Srinivas et~al.(2009)Srinivas, Krause, Kakade, and Seeger}]{Srinivas2009InformationTheoreticRB}
Niranjan Srinivas, Andreas Krause, Sham~M. Kakade, and Matthias~W. Seeger. 2009.
\newblock \href {https://api.semanticscholar.org/CorpusID:59031327} {Information-theoretic regret bounds for gaussian process optimization in the bandit setting}.
\newblock \emph{IEEE Transactions on Information Theory}, 58:3250--3265.

\bibitem[{Team et~al.(2022)Team, Costa-jussà, Cross, Çelebi, Elbayad, Heafield, Heffernan, Kalbassi, Lam, Licht, Maillard, Sun, Wang, Wenzek, Youngblood, Akula, Barrault, Gonzalez, Hansanti, Hoffman, Jarrett, Sadagopan, Rowe, Spruit, Tran, Andrews, Ayan, Bhosale, Edunov, Fan, Gao, Goswami, Guzmán, Koehn, Mourachko, Ropers, Saleem, Schwenk, and Wang}]{nllb2022}
NLLB Team, Marta~R. Costa-jussà, James Cross, Onur Çelebi, Maha Elbayad, Kenneth Heafield, Kevin Heffernan, Elahe Kalbassi, Janice Lam, Daniel Licht, Jean Maillard, Anna Sun, Skyler Wang, Guillaume Wenzek, Al~Youngblood, Bapi Akula, Loic Barrault, Gabriel~Mejia Gonzalez, Prangthip Hansanti, John Hoffman, Semarley Jarrett, Kaushik~Ram Sadagopan, Dirk Rowe, Shannon Spruit, Chau Tran, Pierre Andrews, Necip~Fazil Ayan, Shruti Bhosale, Sergey Edunov, Angela Fan, Cynthia Gao, Vedanuj Goswami, Francisco Guzmán, Philipp Koehn, Alexandre Mourachko, Christophe Ropers, Safiyyah Saleem, Holger Schwenk, and Jeff Wang. 2022.
\newblock \href {http://arxiv.org/abs/2207.04672} {No language left behind: Scaling human-centered machine translation}.

\bibitem[{Trabelsi et~al.(2024)Trabelsi, Vilar, Finkelstein, and Freitag}]{trabelsi2024efficientminimumbayesrisk}
Firas Trabelsi, David Vilar, Mara Finkelstein, and Markus Freitag. 2024.
\newblock \href {http://arxiv.org/abs/2406.02832} {Efficient minimum bayes risk decoding using low-rank matrix completion algorithms}.

\bibitem[{Vamvas and Sennrich(2024)}]{vamvas-sennrich-2024-linear}
Jannis Vamvas and Rico Sennrich. 2024.
\newblock \href {https://doi.org/10.18653/v1/2024.acl-short.71} {Linear-time minimum {B}ayes risk decoding with reference aggregation}.
\newblock In \emph{Proceedings of the 62nd Annual Meeting of the Association for Computational Linguistics (Volume 2: Short Papers)},  790--801, Bangkok, Thailand. Association for Computational Linguistics.

\bibitem[{Vernikos and Popescu-Belis(2024)}]{vernikos-popescu-belis-2024-dont}
Giorgos Vernikos and Andrei Popescu-Belis. 2024.
\newblock \href {https://aclanthology.org/2024.acl-long.653} {Don{'}t rank, combine! combining machine translation hypotheses using quality estimation}.
\newblock In \emph{Proceedings of the 62nd Annual Meeting of the Association for Computational Linguistics (Volume 1: Long Papers)},  12087--12105, Bangkok, Thailand. Association for Computational Linguistics.

\bibitem[{Wang et~al.(2024)Wang, Briakou, Dadkhahi, Agarwal, Cherry, and Cohn}]{wang2024dontthrowawaydata}
Jun Wang, Eleftheria Briakou, Hamid Dadkhahi, Rishabh Agarwal, Colin Cherry, and Trevor Cohn. 2024.
\newblock \href {http://arxiv.org/abs/2407.10456} {Don't throw away data: {Better} sequence knowledge distillation}.

\bibitem[{Wang et~al.(2020)Wang, Wei, Dong, Bao, Yang, and Zhou}]{wang2020minilm}
Wenhui Wang, Furu Wei, Li~Dong, Hangbo Bao, Nan Yang, and Ming Zhou. 2020.
\newblock \href {http://arxiv.org/abs/2002.10957} {{MiniLM}: {Deep} self-attention distillation for task-agnostic compression of pre-trained transformers}.

\bibitem[{Wu et~al.({2020})Wu, Toscano-Palmerin, Frazier, and Wilson}]{pmlr-v115-wu20a}
Jian Wu, Saul Toscano-Palmerin, Peter~I. Frazier, and Andrew~Gordon Wilson. {2020}.
\newblock \href {https://proceedings.mlr.press/v115/wu20a.html} {Practical multi-fidelity bayesian optimization for hyperparameter tuning}.
\newblock In \emph{{Proceedings of The 35th Uncertainty in Artificial Intelligence Conference}}, volume {115} of \emph{{Proceedings of Machine Learning Research}},  {788--798}. PMLR.

\bibitem[{Wu et~al.(2016)Wu, Schuster, Chen, Le, Norouzi, Macherey, Krikun, Cao, Gao, Macherey, Klingner, Shah, Johnson, Liu, Łukasz Kaiser, Gouws, Kato, Kudo, Kazawa, Stevens, Kurian, Patil, Wang, Young, Smith, Riesa, Rudnick, Vinyals, Corrado, Hughes, and Dean}]{wu2016googlesneuralmachinetranslation}
Yonghui Wu, Mike Schuster, Zhifeng Chen, Quoc~V. Le, Mohammad Norouzi, Wolfgang Macherey, Maxim Krikun, Yuan Cao, Qin Gao, Klaus Macherey, Jeff Klingner, Apurva Shah, Melvin Johnson, Xiaobing Liu, Łukasz Kaiser, Stephan Gouws, Yoshikiyo Kato, Taku Kudo, Hideto Kazawa, Keith Stevens, George Kurian, Nishant Patil, Wei Wang, Cliff Young, Jason Smith, Jason Riesa, Alex Rudnick, Oriol Vinyals, Greg Corrado, Macduff Hughes, and Jeffrey Dean. 2016.
\newblock \href {http://arxiv.org/abs/1609.08144} {Google's neural machine translation system: {Bridging} the gap between human and machine translation}.

\bibitem[{Yee et~al.(2019)Yee, Dauphin, and Auli}]{yee-etal-2019-simple}
Kyra Yee, Yann Dauphin, and Michael Auli. 2019.
\newblock \href {https://doi.org/10.18653/v1/D19-1571} {Simple and effective noisy channel modeling for neural machine translation}.
\newblock In \emph{Proceedings of the 2019 Conference on Empirical Methods in Natural Language Processing and the 9th International Joint Conference on Natural Language Processing (EMNLP-IJCNLP)},  5696--5701. Association for Computational Linguistics.

\bibitem[{Zerva et~al.(2022)Zerva, Blain, Rei, Lertvittayakumjorn, C.~de Souza, Eger, Kanojia, Alves, Or{\u{a}}san, Fomicheva, Martins, and Specia}]{zerva-etal-2022-findings}
Chrysoula Zerva, Fr{\'e}d{\'e}ric Blain, Ricardo Rei, Piyawat Lertvittayakumjorn, Jos{\'e}~G. C.~de Souza, Steffen Eger, Diptesh Kanojia, Duarte Alves, Constantin Or{\u{a}}san, Marina Fomicheva, Andr{\'e} F.~T. Martins, and Lucia Specia. 2022.
\newblock \href {https://aclanthology.org/2022.wmt-1.3} {Findings of the {WMT} 2022 shared task on quality estimation}.
\newblock In \emph{Proceedings of the Seventh Conference on Machine Translation (WMT)},  69--99. Association for Computational Linguistics.

\end{thebibliography}
\bibliographystyle{misc/acl_natbib}

\clearpage

\appendix
\onecolumn


    

\begin{table*}[htbp]
\resizebox{\linewidth}{!}{
\begin{tabular}{lccccccccccc}
\toprule
 &  & \multicolumn{10}{c}{\bf CometKiwi runs} \\
\bf Method & \bf Figure & \bf 10 & \bf 20 & \bf 30 & \bf 40 & \bf 50 & \bf 60 & \bf 70 & \bf 80 & \bf 90 & \bf 100 \\
\midrule
UniqRandom & \ref{fig:results_baselines} & 0.7917 & 0.8022 & 0.8074 & 0.8104 & 0.8124 & 0.8140 & 0.8149 & 0.8160 & 0.8168 & 0.8175 \\
LogprobAvg & \ref{fig:results_baselines} & 0.7956 & 0.8055 & 0.8101 & 0.8129 & 0.8149 & 0.8162 & 0.8171 & 0.8181 & 0.8187 & 0.8193 \\
LogprobSum & \ref{fig:results_baselines} & 0.7519 & 0.7723 & 0.7834 & 0.7913 & 0.7974 & 0.8019 & 0.8051 & 0.8081 & 0.8109 & 0.8125 \\
HillClimbing & \ref{fig:results_baselines} & 0.7917 & 0.8080 & 0.8124 & 0.8148 & 0.8165 & 0.8176 & 0.8184 & 0.8191 & 0.8196 & 0.8200 \\
ProxyFirst 200 Distilled-S & \ref{fig:results_multi_SM} & 0.8081 & 0.8141 & 0.8167 & 0.8181 & 0.8190 & 0.8197 & 0.8202 & 0.8206 & 0.8208 & 0.8210 \\
ProxyFirst 200 Distilled-M & \ref{fig:results_multi_SM} & 0.8119 & 0.8165 & 0.8184 & 0.8194 & 0.8201 & 0.8206 & 0.8209 & 0.8211 & 0.8212 & 0.8213 \\
ProxyFirst 50 Distilled-S & \ref{fig:results_multi_SM} & 0.8054 & 0.8100 & 0.8114 & 0.8121 & 0.8124 & - & - & - & - & \\
ProxyFirst 50 Distilled-M & \ref{fig:results_multi_SM} & 0.8073 & 0.8107 & 0.8119 & 0.8122 & 0.8124 & - & - & - & - &  \\
\cmidrule{1-1}
BayesOpt+GP & \ref{fig:results_baselines},\ref{fig:results_multi},\ref{fig:results_multi_SM} & 0.7917 & 0.8121 & 0.8167 & 0.8190 & 0.8201 & 0.8206 & 0.8210 & 0.8212 & 0.8213 & 0.8214 \\
BayesOpt+GP+P with LogprobAvg & \ref{fig:results_multi} & 0.7956 & 0.8123 & 0.8166 & 0.8187 & 0.8198 & 0.8205 & 0.8208 & 0.8210 & 0.8213 & 0.8214 \\
BayesOpt+GP+P with 200 Distilled-S & \ref{fig:results_multi},\ref{fig:results_multi_SM} & 0.8081 & 0.8165 & 0.8190 & 0.8200 & 0.8207 & 0.8210 & 0.8212 & 0.8213 & 0.8214 & 0.8215 \\
BayesOpt+GP+P with 200 Distilled-M & \ref{fig:results_multi},\ref{fig:results_multi_SM} & 0.8119 & 0.8182 & 0.8199 & 0.8205 & 0.8209 & 0.8211 & 0.8213 & 0.8214 & 0.8215 & 0.8215 \\
BayesOpt+GP+P with 50 Distilled-S & \ref{fig:results_multi},\ref{fig:results_multi_SM} & 0.8054 & 0.8153 & 0.8184 & 0.8196 & 0.8204 & 0.8208 & 0.8210 & 0.8213 & 0.8214 & 0.8214 \\
BayesOpt+GP+P with 50 Distilled-M & \ref{fig:results_multi},\ref{fig:results_multi_SM} & 0.8073 & 0.8164 & 0.8187 & 0.8200 & 0.8207 & 0.8209 & 0.8211 & 0.8213 & 0.8214 & 0.8215 \\
\bottomrule\\[1em]
\toprule
 &  & \multicolumn{10}{c}{\bf CometKiwi runs} \\
\bf Method & \bf Figure & \bf 110 & \bf 120 & \bf 130 & \bf 140 & \bf 150 & \bf 160 & \bf 170 & \bf 180 & \bf 190 & \bf 200 \\
\midrule
UniqRandom & \ref{fig:results_baselines} & 0.8182 & 0.8188 & 0.8192 & 0.8197 & 0.8200 & 0.8205 & 0.8208 & 0.8211 & 0.8214 & 0.8216 \\
LogprobAvg & \ref{fig:results_baselines} & 0.8199 & 0.8203 & 0.8205 & 0.8209 & 0.8211 & 0.8212 & 0.8213 & 0.8214 & 0.8216 & 0.8216 \\
LogprobSum & \ref{fig:results_baselines} & 0.8139 & 0.8156 & 0.8170 & 0.8180 & 0.8188 & 0.8196 & 0.8204 & 0.8209 & 0.8212 & 0.8216 \\
HillClimbing & \ref{fig:results_baselines} & 0.8203 & 0.8206 & 0.8208 & 0.8209 & 0.8211 & 0.8213 & 0.8214 & 0.8215 & 0.8216 & 0.8216 \\
\cmidrule{1-1}
BayesOpt+GP & \ref{fig:results_baselines},\ref{fig:results_multi},\ref{fig:results_multi_SM} & 0.8215 & 0.8215 & 0.8215 & 0.8216 & 0.8216 & 0.8216 & 0.8216 & 0.8216 & 0.8216 & 0.8216 \\
BayesOpt+GP+P with LogprobAvg & \ref{fig:results_multi} & 0.8214 & 0.8215 & 0.8215 & 0.8216 & 0.8216 & 0.8216 & 0.8216 & 0.8216 & 0.8216 & 0.8216 \\
\bottomrule
\end{tabular}
}
\caption{Exact values (selected candidate score) for \Cref{fig:results_baselines,fig:results_multi,fig:results_multi_SM}.}
\label{tab:results_exact_values}
\end{table*}

\clearpage

\section{Statistical Significance}
\label{sec:statistical_significance}

We measure statistical significance between two methods based on the final candidate \cometkiwi scores with either budget 30, 60, 90, or across the budget range from 10 to 190 in \Cref{tab:statistical_head_to_head}.
To determine whether one method is better than another one, we use one-sided paired Student's t-test with p-value threshold $0.01$ which is run across the individual samples.

\begin{table}[htbp]
\resizebox{0.49\linewidth}{!}{
\begin{tabular}{rcccccc}
\bf \Large Budget 30 \hspace{2cm} & \rotatebox{90}{UniqRandom} & \rotatebox{90}{LogprobAvg} & \rotatebox{90}{LogprobSum} & \rotatebox{90}{HillClimbing} & \rotatebox{90}{BayesOpt+GP} \\
\midrule
                                   LogprobAvg & $\leftarrow$ &              & $\leftarrow$ & $\uparrow$   & $\uparrow$   & \\
                                   LogprobSum & $\uparrow$   & $\uparrow$   &              & $\uparrow$   & $\uparrow$   & \\
                                 HillClimbing & $\leftarrow$ & $\leftarrow$ & $\leftarrow$ &              & $\uparrow$   & \\
                   ProxyFirst 200 Distilled-S & $\leftarrow$ & $\leftarrow$ & $\leftarrow$ & $\leftarrow$ &              & \\
                   ProxyFirst 200 Distilled-M & $\leftarrow$ & $\leftarrow$ & $\leftarrow$ & $\leftarrow$ & $\leftarrow$ & \\
                    ProxyFirst 50 Distilled-S & $\leftarrow$ & $\leftarrow$ & $\leftarrow$ & $\uparrow$   & $\uparrow$   & \\
                    ProxyFirst 50 Distilled-M & $\leftarrow$ & $\leftarrow$ & $\leftarrow$ &              & $\uparrow$   & \\
\midrule
                                  BayesOpt+GP & $\leftarrow$ & $\leftarrow$ & $\leftarrow$ & $\leftarrow$ &              & \\
                  BayesOpt+GP+P with LogprobAvg & $\leftarrow$ & $\leftarrow$ & $\leftarrow$ & $\leftarrow$ &              & \\
             BayesOpt+GP+P with 200 Distilled-S & $\leftarrow$ & $\leftarrow$ & $\leftarrow$ & $\leftarrow$ & $\leftarrow$ & \\
             BayesOpt+GP+P with 200 Distilled-M & $\leftarrow$ & $\leftarrow$ & $\leftarrow$ & $\leftarrow$ & $\leftarrow$ & \\
              BayesOpt+GP+P with 50 Distilled-S & $\leftarrow$ & $\leftarrow$ & $\leftarrow$ & $\leftarrow$ & $\leftarrow$ & \\
              BayesOpt+GP+P with 50 Distilled-M & $\leftarrow$ & $\leftarrow$ & $\leftarrow$ & $\leftarrow$ & $\leftarrow$ & \\
\bottomrule
\end{tabular}
}
\resizebox{0.49\linewidth}{!}{
\begin{tabular}{rcccccc}
\bf \Large Budget 60 \hspace{2cm} & \rotatebox{90}{UniqRandom} & \rotatebox{90}{LogprobAvg} & \rotatebox{90}{LogprobSum} & \rotatebox{90}{HillClimbing} & \rotatebox{90}{BayesOpt+GP} \\
\midrule
                                   LogprobAvg & $\leftarrow$ &              & $\leftarrow$ & $\uparrow$   & $\uparrow$   & \\
                                   LogprobSum & $\uparrow$   & $\uparrow$   &              & $\uparrow$   & $\uparrow$   & \\
                                 HillClimbing & $\leftarrow$ & $\leftarrow$ & $\leftarrow$ &              & $\uparrow$   & \\
                   ProxyFirst 200 Distilled-S & $\leftarrow$ & $\leftarrow$ & $\leftarrow$ & $\leftarrow$ & $\uparrow$   & \\
                   ProxyFirst 200 Distilled-M & $\leftarrow$ & $\leftarrow$ & $\leftarrow$ & $\leftarrow$ &              & \\

                   \\
                   \\
\midrule
                                  BayesOpt+GP & $\leftarrow$ & $\leftarrow$ & $\leftarrow$ & $\leftarrow$ &              & \\
                  BayesOpt+GP+P with LogprobAvg & $\leftarrow$ & $\leftarrow$ & $\leftarrow$ & $\leftarrow$ &              & \\
             BayesOpt+GP+P with 200 Distilled-S & $\leftarrow$ & $\leftarrow$ & $\leftarrow$ & $\leftarrow$ & $\leftarrow$ & \\
             BayesOpt+GP+P with 200 Distilled-M & $\leftarrow$ & $\leftarrow$ & $\leftarrow$ & $\leftarrow$ & $\leftarrow$ & \\
              BayesOpt+GP+P with 50 Distilled-S & $\leftarrow$ & $\leftarrow$ & $\leftarrow$ & $\leftarrow$ &              & \\
              BayesOpt+GP+P with 50 Distilled-M & $\leftarrow$ & $\leftarrow$ & $\leftarrow$ & $\leftarrow$ & $\leftarrow$ & \\
\bottomrule
\end{tabular}
}

\bigskip
\bigskip

\resizebox{0.49\linewidth}{!}{
\begin{tabular}{rcccccc}
\bf \Large Budget 90 \hspace{2cm} & \rotatebox{90}{UniqRandom} & \rotatebox{90}{LogprobAvg} & \rotatebox{90}{LogprobSum} & \rotatebox{90}{HillClimbing} & \rotatebox{90}{BayesOpt+GP} \\
\midrule
                                   LogprobAvg & $\leftarrow$ &              & $\leftarrow$ & $\uparrow$   & $\uparrow$   & \\
                                   LogprobSum & $\uparrow$   & $\uparrow$   &              & $\uparrow$   & $\uparrow$   & \\
                                 HillClimbing & $\leftarrow$ & $\leftarrow$ & $\leftarrow$ &              & $\uparrow$   & \\
                   ProxyFirst 200 Distilled-S & $\leftarrow$ & $\leftarrow$ & $\leftarrow$ & $\leftarrow$ & $\uparrow$   & \\
                   ProxyFirst 200 Distilled-M & $\leftarrow$ & $\leftarrow$ & $\leftarrow$ & $\leftarrow$ &              & \\
\midrule
                                  BayesOpt+GP+P & $\leftarrow$ & $\leftarrow$ & $\leftarrow$ & $\leftarrow$ &              & \\
                  BayesOpt+GP+P with LogprobAvg & $\leftarrow$ & $\leftarrow$ & $\leftarrow$ & $\leftarrow$ &              & \\
             BayesOpt+GP+P with 200 Distilled-S & $\leftarrow$ & $\leftarrow$ & $\leftarrow$ & $\leftarrow$ &              & \\
             BayesOpt+GP+P with 200 Distilled-M & $\leftarrow$ & $\leftarrow$ & $\leftarrow$ & $\leftarrow$ & $\leftarrow$ & \\
              BayesOpt+GP+P with 50 Distilled-S & $\leftarrow$ & $\leftarrow$ & $\leftarrow$ & $\leftarrow$ &              & \\
              BayesOpt+GP+P with 50 Distilled-M & $\leftarrow$ & $\leftarrow$ & $\leftarrow$ & $\leftarrow$ &              & \\
\bottomrule
\end{tabular}
}
\resizebox{0.49\linewidth}{!}{
\begin{tabular}{rcccccc}
\bf \Large Across budgets 10 to 190 \hspace{-0.2cm} & \rotatebox{90}{UniqRandom} & \rotatebox{90}{LogprobAvg} & \rotatebox{90}{LogprobSum} & \rotatebox{90}{HillClimbing} & \rotatebox{90}{BayesOpt+GP} \\
\midrule
                                   LogprobAvg & $\leftarrow$ &              & $\leftarrow$ & $\uparrow$   & $\uparrow$   & \\
                                   LogprobSum & $\uparrow$   & $\uparrow$   &              & $\uparrow$   & $\uparrow$   & \\
                                 HillClimbing & $\leftarrow$ & $\leftarrow$ & $\leftarrow$ &              & $\uparrow$   & \\
                   ProxyFirst 200 Distilled-S & $\leftarrow$ & $\leftarrow$ & $\leftarrow$ & $\leftarrow$ & $\leftarrow$ & \\
                   ProxyFirst 200 Distilled-M & $\leftarrow$ & $\leftarrow$ & $\leftarrow$ & $\leftarrow$ & $\leftarrow$ & \\
\midrule
                                  BayesOpt+GP & $\leftarrow$ & $\leftarrow$ & $\leftarrow$ & $\leftarrow$ &              & \\
                  BayesOpt+GP+P with LogprobAvg & $\leftarrow$ & $\leftarrow$ & $\leftarrow$ & $\leftarrow$ &              & \\
             BayesOpt+GP+P with 200 Distilled-S & $\leftarrow$ & $\leftarrow$ & $\leftarrow$ & $\leftarrow$ & $\leftarrow$ & \\
             BayesOpt+GP+P with 200 Distilled-M & $\leftarrow$ & $\leftarrow$ & $\leftarrow$ & $\leftarrow$ & $\leftarrow$ & \\
              BayesOpt+GP+P with 50 Distilled-S & $\leftarrow$ & $\leftarrow$ & $\leftarrow$ & $\leftarrow$ & $\leftarrow$ & \\
              BayesOpt+GP+P with 50 Distilled-M & $\leftarrow$ & $\leftarrow$ & $\leftarrow$ & $\leftarrow$ & $\leftarrow$ & \\
\bottomrule
\end{tabular}
}
\caption{
Statistical significance comparison between proposed methods across various \cometkiwi calls budgets.
Within a cell, $\uparrow$ means that the column method (in header) is statistically significantly better than the row method and $\leftarrow$ means the opposite.
If a cell is empty, none of the methods are significantly better than the other one.
For example, in Budget 30 (top left) table, in third row and first column, $\leftarrow$ means that HillClimbing is significantly better than UniqRandom in the setup of budget of 30.
}
\label{tab:statistical_head_to_head}
\end{table}

\end{document}